\documentclass{article}

\usepackage{arxiv}

\usepackage[utf8]{inputenc} 
\usepackage[T1]{fontenc}    
\usepackage{hyperref}       
\usepackage{url}            
\usepackage{booktabs}       
\usepackage{amsfonts}       
\usepackage{nicefrac}       
\usepackage{microtype}      
\usepackage{lipsum}
\usepackage{graphicx}
\graphicspath{ {./images/} }

\title{So You Need Datasets for Your COVID-19 Detection Research Using Machine Learning?}

\author{
 Md Fahimuzzman Sohan \\
  Department of Software Engineering \\
  Daffodil International University\\
  Dhaka 1207, Bangladesh \\
  \texttt{fahimsohan2@gmail.com} \\
}

\begin{document}
\maketitle
\begin{abstract}
Millions of people are infected by the coronavirus disease 2019 (COVID‐19) around the world. Machine Learning (ML) techniques are being used for COVID‐19 detection research from the beginning of the epidemic. This article represents the detailed information on frequently used datasets in COVID-19 detection using Machine Learning (ML). We investigated 96 papers on COVID-19 detection between January 2020 and June 2020. We extracted the information about used datasets from the articles and represented them here simultaneously. This investigation will help future researchers to find the COVID-19 datasets without difficulty.  
\end{abstract}


\section{Introduction}
The Severe Acute Respiratory Syndrome Coronavirus 2, also known as SARS-CoV-2 or novel Coronavirus 2019 or mostly used COVID-19 was first reported in December 2019 in the Hubei Province of China \cite{rodriguez2020clinical}. Over 0.7 million deaths and around 20 million confirmed cases had been reported worldwide by the World Health Organization (WHO) due to the virus epidemic \cite{worldhealthorganization}. A vital step is quickly identifying the infected people and isolating them to delay the spread of the epidemic \cite{wynants2020prediction}. In the context of COVID-19 identification, ML techniques have been employed to detect the disease from various data analysis and classification \cite{elaziz2020new,brinati2020detection}. ML-based techniques are automatic and easy to use and implement in clinical settings \cite{khuzani2020covid}. One of the key parts of ML techniques is the dataset to train detection models and validation. Literature shows, most of the COVID-19 detection-based models were developed using image datasets including X-ray image, Computed Tomography (CT) images. But it is difficult and hard work for new researchers to collect and identify the required dataset for their research. 

To address this issue, we represented a large scale open access datasets collection which was used by previous COVID-19 research studies. Initially, we collected hundreds of COVID-19 detection related articles between January 2020 and June 2020 from various online libraries, such as Google Scholar, Elsevier, PubMed, and WHO Database. Finally, we selected 96 ML-based COVID-19 detection articles to conduct this investigation. This article collection process was performed by the established and popular Preferred Reporting Items for Systematic Reviews and Meta-Analyses (PRISMA) statement \cite{moher2009preferred}. Then we extracted the information of used datasets for detection models from each article. We listed out the most frequently used datasets in the 96 studies. Besides, we included the detailed information, open accessible destination link, and available paper link of each dataset in the next section. This dataset compilation initiative will give easy and effective access for future COVID-19 researchers.

\section{Task description}

\begin{figure}[h]
\centering
\includegraphics[width=0.90\linewidth]{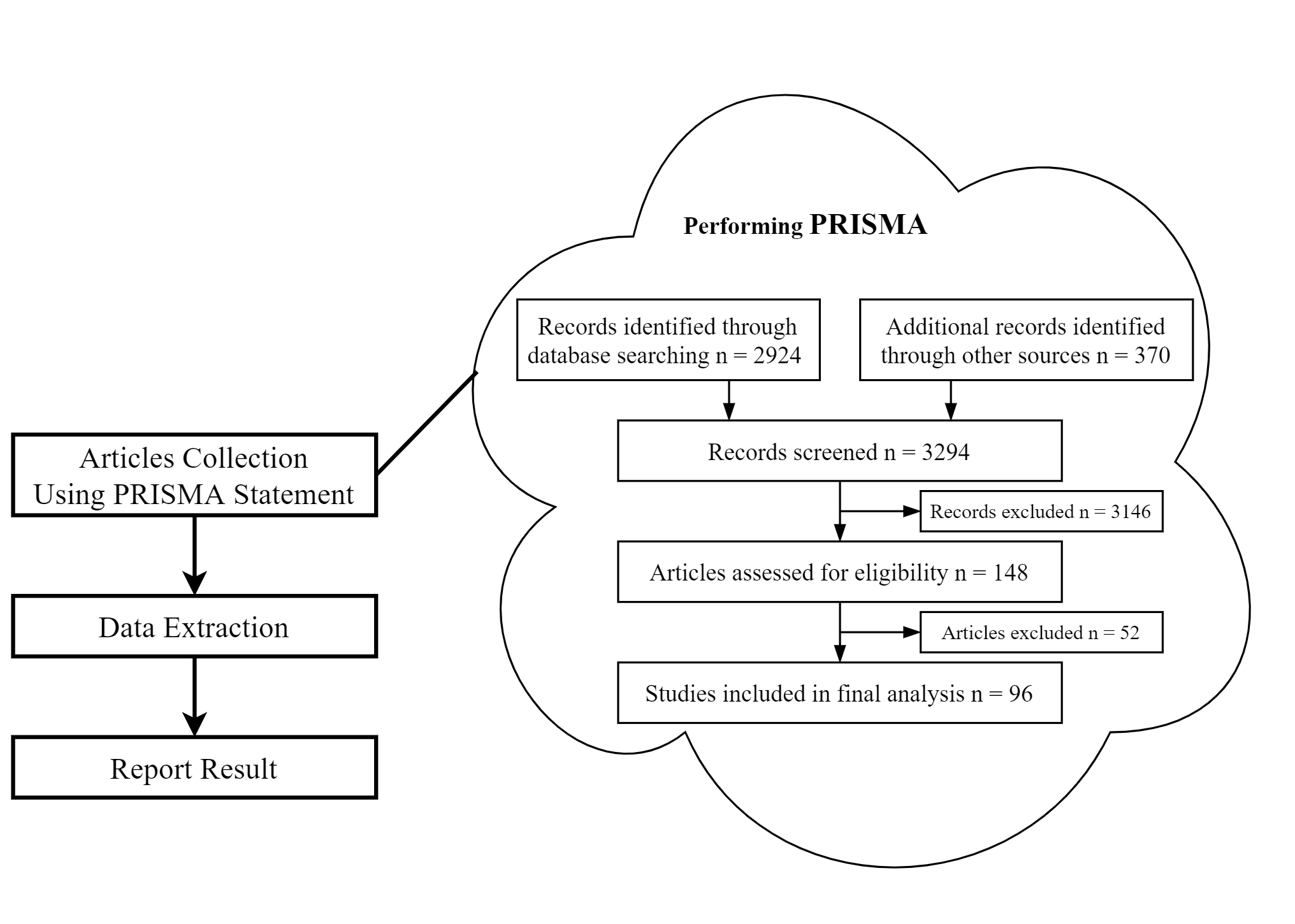}
\caption{The steps of conducting this investigation, here PRISMA Statement was used only for the collecting required articles}
\label{fig:1}
\end{figure}

As Figure \ref{fig:1} shows, the first task of this study was collecting articles from different databases. We used the PRISMA statement to collect all articles that are relevant to COVID-19 detection research using ML techniques. The article collection using the PRISMA method was described broadly in the following articles \cite{rodriguez2020clinical,wynants2020prediction,salehi2020coronavirus}. In addition, we used four digital libraries to collect articles including Google Scholar, Elsevier, PubMed, and WHO Database. After ending this process we confirmed 96 COVID-19 detection research articles using ML techniques. The next step was extraction data from the 96 collected studies. We collected the information on the used dataset from each article individually. The considered articles list and extracted datasets information are available at \url{https://figshare.com/s/0cee05e6405a9310802e}. Finally, the extracted data were summarised and the result presented in the next section.

\section{Resulting Datasets}

\begin{table}[]
\caption{List of most used datasets in the available COVID-19 detection research papers using ML}
\centering
\begin{tabular}{llc}
\hline
\textbf{No.} & \textbf{Name} & \multicolumn{1}{l}{\textbf{No. of Article Used}} \\ \hline
1 & Cohen   JP dataset & 60 \\
2 & Chest   X-Ray Images Pneumonia & 26 \\
3 & RSNA   Dataset & 14 \\
4 & COVID-CT & 13 \\
5 & ChestX-ray8 & 11 \\
6 & Covid-19-database & 10 \\
7 & COVIDx & 9 \\
8 & Mendeley   Dataset & 9 \\
9 & Qatar-Dhaka   COVID-19 Data & 7 \\
10 & NIH   Dataset & 5 \\
11 & Twitter   Data & 5 \\
12 & Snapshots   data & 4 \\
13 & COVID-CTset & 1 \\
14 & IRCCS blood Data & 1 \\
15 & POCUS Dataset & 1 \\ \hline
\end{tabular}
\label{tab1}
\end{table}

At the beginning of the discussion of the datasets, we represented commonly used 15 datasets with respective dataset names and the number of articles used the dataset in Table \ref{tab1}. Moreover, we divided the datasets into three categories by the appearance of COVID-19 positive and negative samples. The three categories are COVID-19, Non-COVID-19, and lastly COVID-19 and non-COVID-19 Data. In COVID-19 datasets, only COVID-19 positive cases were included, negative COVID-19 cases were included in the non-COVID-19 dataset, and COVID-19 and non-COVID-19 datasets contain both positive and negative cases. In the following subsections and tables, we represented the three categories datasets elaborately including a short description, characteristics, and reference link of each dataset. 

\subsection{COVID-19 Data (Table \ref{tab2})}

\begin{table}[]
\centering
\caption{List of COVID-19 positive dataset with data characteristics and open access data links}
\begin{tabular}{llll}
\hline
\textbf{No.} & \textbf{Name} & \textbf{Characteristics} & \textbf{Reference} \\ \hline
01 & \begin{tabular}[c]{@{}l@{}}Covid-19\\ Database (Accessed\\ 22-Jul-2020)\end{tabular} & \begin{tabular}[c]{@{}l@{}} $\bullet$ Chest CT images\\ $\bullet$ 68 cases\end{tabular} & \begin{tabular}[c]{@{}l@{}} $\bullet$ \href{https://www.sirm.org/en/category/articles/covid-19-database/}{Link to data} \end{tabular} \\ \hline
02 & \begin{tabular}[c]{@{}l@{}}Twitter Data\\ (Accessed 22-Jul-2020)\end{tabular} & \begin{tabular}[c]{@{}l@{}} $\bullet$ Chest X-ray images\\ $\bullet$ Total 135 images of 50 cases (last access 22-Jul-2020)\\$\bullet$ JFIF format at approximately a 2K$\times$2K resolution \end{tabular} & \begin{tabular}[c]{@{}l@{}} $\bullet$ \href{https://twitter.com/ChestImaging}{Link to data} \end{tabular} \\ \hline
03 & \begin{tabular}[c]{@{}l@{}}Snapshots Data \\ (Accessed 22-Jul-2020)\end{tabular} & \begin{tabular}[c]{@{}l@{}} $\bullet$ Chest X-ray images\\ $\bullet$ Total 73 confirmed COVID-19 cases (last access 22-Jul-2020) \end{tabular} & \begin{tabular}[c]{@{}l@{}} $\bullet$ \href{https://www.kaggle.com/andrewmvd/convid19-X-rays}{Link to data} \end{tabular}  \\
\hline
\end{tabular}
\label{tab2}
\end{table}

\begin{enumerate}
    \item \textbf{Covid-19 Database:} It is a public database of COVID-19 chest CT images by the Italian Society of Medical and Interventional Radiology (SIRM). They have continuously uploaded the images on the website. Every image contains a patient's metadata, such as Age, Gender, and relevant symptom histories.
    \item \textbf{Twitter Data:} This image data are available on Twitter, where a cardiothoracic radiologist from Spain has shared high-quality COVID-19 positive subjects.
    \item \textbf{Snapshots Data:} This dataset was prepared by collecting data from RSNA, Radiopedia, and COVID-19-database Dataset. This dataset is publicly available in the Kaggle repository.
\end{enumerate}

\subsection{Non-COVID-19 Data (Table \ref{tab3})}

\begin{enumerate}
  \item \textbf{ChestX-ray8} This dataset is prepared by a large number of chest X-ray images of several lung diseases and known as “ChestX-ray8”. The data were collected between the year 1992 to 2015 and from various platforms. This dataset is publicly available for research purposes and reported as most commonly accessible in medical image investigation. In COVID-19 research, mostly researchers used a portion of this large dataset.
  \item \textbf{NIH Dataset:} It’s a subset of the “ChestX-ray8” dataset. This sample dataset contains 5\% of the full version. This dataset was published under the National Institutes of Health (NIH), USA.
  \item \textbf{Mendeley Dataset:} This large scale non-COVID dataset was published in February 2018. The data were collected from a Children’s medical center in Guangzhou, China. This repository of images is made available freely for research purposes. 
  \item \textbf{Chest X-Ray Images Pneumonia:} This chest X-ray images dataset is known as “Chest X-Ray Images Pneumonia”, a part of Mendeley and Cohen JP Dataset. Authority prepared the dataset by screening and checking raw images to ensure quality. The dataset is available online with open access. 
  \item \textbf{RSNA Dataset:} This is the second version of the dataset, where more data is added with the previous version. It is developed jointly by the Radiological Society of North America, US National Institutes of Health, The Society of Thoracic Radiology, and MD.ai. This dataset is also a part of “ChestX-ray8” and updated continuously.
\end{enumerate}

\begin{table}[]
\centering
\caption{List of Non-COVID-19 datasets with data characteristics and freely accessible data links}
\begin{tabular}{llll}
\hline
\textbf{No.} & \textbf{Name} & \textbf{Characteristics} & \textbf{Reference} \\ \hline
01 & \begin{tabular}[c]{@{}l@{}} ChestX-ray8\\ (Accessed\\ 10-Jul-2020) \end{tabular} & \begin{tabular}[c]{@{}l@{}} $\bullet$ Large number of chest X-ray images\\ $\bullet$ 14 different lung diseases and normal images \\ $\bullet$ Total 108,948 frontal-view X-ray images\\ $\bullet$ 11 metadata\end{tabular} & \begin{tabular}[c]{@{}l@{}} $\bullet$ \href{https://nihcc.app.box.com/v/ChestXray-NIHCC/file/256057377774}{Link to data}\\ $\bullet$ \href{https://openaccess.thecvf.com/content_cvpr_2017/html/Wang_ChestX-ray8_Hospital-Scale_Chest_CVPR_2017_paper.html}{Link to paper} \end{tabular} \\ \hline
02 & \begin{tabular}[c]{@{}l@{}} NIH Dataset\\ (Accessed\\ 10-Jul-2020) \end{tabular} & \begin{tabular}[c]{@{}l@{}} $\bullet$ Chest X-ray images\\ $\bullet$ 14 different lung diseases and normal images  \\ $\bullet$ 5,606 images with size 1024x1024\\ $\bullet$ 11 metadata\end{tabular} & \begin{tabular}[c]{@{}l@{}} $\bullet$ \href{https://www.kaggle.com/nih-chest-xrays/sample}{Link to data}\\ $\bullet$ \href{https://www.nih.gov/news-events/news-releases/nih-clinical-center-provides-one-largest-publicly-available-chest-x-ray-datasets-scientific-community}{Link to paper} \end{tabular} \\ \hline
03 & \begin{tabular}[c]{@{}l@{}}Mendeley\\ Dataset\\ (Accessed\\ 10-Jul-2020)\end{tabular} & \begin{tabular}[c]{@{}l@{}} $\bullet$ Optical Coherence Tomography (OCT) and Chest X-Ray Images \\ $\bullet$ 108,312 OCT images (four types: CNV, DME, DRUSEN, and\\ NORMAL)\\ $\bullet$ 5,232 chest X-ray images (3,883 pneumonia and 1,349 normal)\end{tabular} & \begin{tabular}[c]{@{}l@{}} $\bullet$ \href{https://data.mendeley.com/datasets/rscbjbr9sj/2}{Link to data}\\ $\bullet$ \href{https://www.cell.com/cell/fulltext/S0092-8674(18)30154-5}{Link to paper} \end{tabular} \\ \hline
04 & \begin{tabular}[c]{@{}l@{}}Chest X-Ray\\ Images Pneumonia\\ (Accessed\\ 11-Jul-2020)\end{tabular} & \begin{tabular}[c]{@{}l@{}} $\bullet$ Chest X-ray images\\ $\bullet$ Two types images: pneumonia and normal\\ $\bullet$ Total 5,856 images, 4273 pneumonia and 1583 normal\\ $\bullet$ JPEG file format with  around 1168x984 pixels\end{tabular} & $\bullet$  \href{https://www.kaggle.com/paultimothymooney/chest-xray-pneumonia?}{Link to data} \\ \hline
05 & \begin{tabular}[c]{@{}l@{}}RSNA Dataset\\ (Accessed\\ 14-Jul-2020)\end{tabular} & \begin{tabular}[c]{@{}l@{}} $\bullet$ Chest X-ray images\\ $\bullet$ Total 30227 samples found; 8851 normal and 21376 infected\\ $\bullet$ DICOM file format\end{tabular} & $\bullet$ \href{https://www.kaggle.com/c/rsna-pneumonia-detection-challenge}{Link to data} \\ \hline
\end{tabular}
\label{tab3}
\end{table}

\subsection{COVID-19 and non-COVID-19 Data (Table \ref{tab4})}

\begin{table}[]
\centering
\caption{List of OVID-19 and non-COVID-19 dataset with data characteristics and open access data links}
\begin{tabular}{llll}
\hline
\textbf{No.} & \textbf{Name} & \textbf{Characteristics} & \textbf{Reference} \\ \hline
01 & \begin{tabular}[c]{@{}l@{}}Cohen JP\\ dataset\\ (Accessed\\ 09-Jul-2020)\end{tabular} & \begin{tabular}[c]{@{}l@{}} $\bullet$ Chest X-ray and CT images\\ $\bullet$ Various types of pneumonia were applied including viral, bacterial,\\ fungal, and lipoid\\ $\bullet$ Total 589 images; 542 are frontal and 47 are lateral view\\ $\bullet$ 434 COVID-19 positive images\\ $\bullet$ 16 metadata\\ $\bullet$ Approved by the University of Montreal’s Ethics Committee\end{tabular} & \begin{tabular}[c]{@{}l@{}} $\bullet$ \href{https://github.com/ieee8023/covid-chestxray-dataset}{Link to data} \\ $\bullet$ \href{https://arxiv.org/abs/2006.11988}{Link to paper} \\ \end{tabular} \\ \hline
02 & \begin{tabular}[c]{@{}l@{}}Qatar-Dhaka\\ COVID-19\\ Data (Accessed\\ 10-Jul-2020)\end{tabular} & \begin{tabular}[c]{@{}l@{}} $\bullet$ Chest X-ray images\\ $\bullet$ COVID-19, pneumonia, and normal images were included\\ $\bullet$ Total 2905 images, including 219 COVID-19 positive, 1345 viral\\ pneumonia, and 1341 normal images\\ $\bullet$ Portable Network Graphics (PNG) file format with 1024-by-1024\\ pixels resolution \end{tabular} & \begin{tabular}[c]{@{}l@{}} $\bullet$ \href{https://www.kaggle.com/tawsifurrahman/covid19-radiography-database}{Link to data}\\ $\bullet$ \href{https://arxiv.org/abs/2003.13145}{Link to paper}\end{tabular} \\ \hline
03 & \begin{tabular}[c]{@{}l@{}}COVIDx\\ (Accessed\\ 11-Jul-2020)\end{tabular}  & \begin{tabular}[c]{@{}l@{}} $\bullet$ Chest X-ray images\\ $\bullet$ Three types images: COVID-19, pneumonia, and normal\\ $\bullet$ Total 14198 images; 8066 normal, 5559 pneumonia, and 573\\ COVID-19\end{tabular} & \begin{tabular}[c]{@{}l@{}} $\bullet$ \href{https://github.com/lindawangg/COVID-Net}{Link to data}\\ $\bullet$ \href{https://arxiv.org/abs/2003.09871}{Link to paper}\end{tabular} \\ \hline
04 & \begin{tabular}[c]{@{}l@{}}COVID-CT\\ (Accessed\\ 12-Jul-2020)\end{tabular} & \begin{tabular}[c]{@{}l@{}}$\bullet$ Chest CT images\\ $\bullet$ COVID-CT contains 349 images  from 216 patients non-COVID-19\\ CT contains 463 images from 55 patients\\ $\bullet$ More than seven metadata\end{tabular} & \begin{tabular}[c]{@{}l@{}} $\bullet$ \href{https://github.com/UCSD-AI4H/COVID-CT}{Link to data}\\ $\bullet$ \href{https://arxiv.org/abs/2003.13865}{Link to paper}\end{tabular} \\ \hline
05 & \begin{tabular}[c]{@{}l@{}}COVID-CTset\\ (Accessed\\ 13-Jul-2020)\end{tabular} & \begin{tabular}[c]{@{}l@{}} $\bullet$ Lung CT scan images\\ $\bullet$ Total 63849 images; 48260 normal and 15589 COVIT-19 positive\\ $\bullet$ Four metadata\\ $\bullet$ Tagged Image File Format (TIFF)\end{tabular} & \begin{tabular}[c]{@{}l@{}} $\bullet$ \href{https://github.com/mr7495/COVID-CTset}{Link to data}\\ $\bullet$ \href{https://www.medrxiv.org/content/10.1101/2020.06.08.20121541v2}{Link to paper}\end{tabular} \\ \hline
06 & \begin{tabular}[c]{@{}l@{}}POCUS Dataset\\ (Accessed\\ 14-Jul-2020)\end{tabular} & \begin{tabular}[c]{@{}l@{}} $\bullet$ Lung ultrasound images\\ $\bullet$ Collected from 64 lung POCUS video recordings\\ $\bullet$ Total 1103 images; 654 COVID-19, 277 pneumonia, and 172 normal\\ $\bullet$ 60\% COVID-19 data \\ $\bullet$ Seven metadata \end{tabular} & \begin{tabular}[c]{@{}l@{}} $\bullet$ \href{https://github.com/jannisborn/covid19_pocus_ultrasound}{Link to data}\\ $\bullet$ \href{https://arxiv.org/abs/2004.12084}{Link to paper}\end{tabular} \\ \hline
07 & \begin{tabular}[c]{@{}l@{}}IRCCS blood\\ Data (Accessed\\ 14-Jul-2020)\end{tabular} & \begin{tabular}[c]{@{}l@{}} $\bullet$ Routine blood exam records\\ $\bullet$ 16 metadata \\ $\bullet$ Total 279 cases; 177 Non COVID-19 and 102 COVID-19 \\ $\bullet$ Numerical (continuous) data type except Gender, Age, and Swab data\end{tabular} & \begin{tabular}[c]{@{}l@{}} $\bullet$ \href{https://zenodo.org/record/3886927#.Xxhnt54zaUm}{Link to data}\\ $\bullet$ \href{https://www.medrxiv.org/content/10.1101/2020.04.22.20075143v1}{Link to paper}\end{tabular}

\\ \hline
\end{tabular}
\label{tab4}
\end{table}

\begin{enumerate}
  \item \textbf{Cohen JP Dataset:} This dataset is known as the “Cohen JP dataset”, widely used, and reported as the very first COVID-19 image dataset. This open-access dataset contains chest X-ray and CT images collected from different hospitals of different countries. All data is released under the GitHub repository and updated continuously by the authority.
  \item \textbf{Qatar-Dhaka COVID-19 Data:} It is prepared under the collaboration of some researchers from Bangladesh, Qatar, Pakistan, and Malaysia; we named it as “Qatar-Dhaka COVID-19 Data”. Data collected from different sources are combined in their dataset, e.g., Cohen JP dataset, COVID-19-database. This combined dataset is available online and freely accessible in the Kaggle repository.
  \item \textbf{COVIDx:} It is also a hybrid dataset that combines 5 different publicly available data repositories and is known as “COVIDx”. This open-access dataset is available in GitHub storage. Three types of data were included and each type has different train and test data distribution in the dataset.
  \item \textbf{COVID-CT:} This dataset is divided into two parts: one for COVIT-19 positive known as “COVID-CT” and non-COVIT-19 known as “non-COVID-19 CT”. Data of various sources were combined to prepare the datasets. The contributors collected 760 COVID-19 research preprints from two different platforms. Then they did an interesting job, they extracted CT images from the PDF documents of the articles. Extracted images were manually pre-processed and metadata of each image collected. 
  \item \textbf{COVID-CTset:} The lung CT scan data was collected from an Iranian medical center. Dataset contains COVID-19 positive and normal both patients' images, also they included the metadata of the patients. The data and relevant documents are available online with open access.
  \item \textbf{POCUS Dataset:} This dataset was prepared differently, images were taken from lung POCUS video recordings that are publicly available in web and publications. A total of 64 videos were considered from various sources, where 39 COVID-19, 14 pneumonia, and 11 videos of healthy patients. Average 17$\pm$6 frames were selected from per video with a frame rate of 3Hz. Dataset is available in the GitHub repository with open access.
  \item \textbf{IRCCS blood data:} The data of the dataset were collected from patients' blood, known as “IRCCS blood data”. This dataset consists of routine blood exam records of patients admitted to an Italian hospital from the end of February 2020 to mid of March 2020.  Several features were considered from blood tests, including white blood cell counts, and the platelets, CRP, AST, ALT, GGT, ALP, LDH plasma levels. They also included the rRT-PCR swab test results for every patient as the dependent variable. Dataset is freely available for everyone in the Zenodo repository.
\end{enumerate}

\newpage
\section{Conclusion}

We build a collection of COVID-19 research datasets, where top used datasets were considered from the COVID-19 detection research articles between the period of January 2020 and June 2020. We investigated 96 research articles to access the information of used datasets. We described the datasets in three categories, one was COVID-19 positive sample-based, another was negative sample-based, and lastly, positive-negative both samples based. We represented the datasets along with a short description, Characteristics of them, and relevant reference links. This article will help future researchers to find the COVID-19 detection research datasets easily and it will save their time.

\bibliographystyle{unsrt}  


\end{document}